\documentclass[conference]{IEEEtran}
\IEEEoverridecommandlockouts

\usepackage{amsmath,amssymb,amsfonts}
\usepackage{graphicx}
\usepackage{textcomp}
\usepackage{algorithm}
\usepackage{algpseudocode}
\usepackage{xcolor}
\usepackage{tikz}
\usetikzlibrary{positioning,arrows.meta}
\usepackage{pgfplots}
\pgfplotsset{compat=1.17}
\usepackage{hyperref}
\hypersetup{
  citebordercolor={0 0.6 0},
  linkbordercolor={1 1 1},
  urlbordercolor={1 1 1}
}
\def\BibTeX{{\rm B\kern-.05em{\sc i\kern-.025em b}\kern-.08em
    T\kern-.1667em\lower.7ex\hbox{E}\kern-.125emX}}
\begin{document}

\title{MemDecay: Region-Aware KV Cache Eviction for Efficient LLM Agent Inference}

\author{\IEEEauthorblockN{Venkatesha Matam}
\IEEEauthorblockA{\textit{Independent Researcher} \\
New York City, USA \\
matam.venkatesha@gmail.com}
\and
\IEEEauthorblockN{Keon Kim}
\IEEEauthorblockA{\textit{Independent Researcher} \\
New York City, USA \\
kwk236@gmail.com}
}

\maketitle

\begin{abstract}
Large language models (LLMs) are increasingly deployed as agents whose contexts accumulate system instructions, plans, user turns, tool outputs, retrieved documents, and intermediate reasoning over many steps. The key--value (KV) cache that supports autoregressive attention grows with this trace, and its memory footprint constrains long-horizon inference and serving. Existing KV-cache eviction policies score tokens with attention or recency signals and apply the same retention rule to the entire sequence, leaving the semantic structure of orchestrated agent prompts unused.

We propose \textbf{MemDecay}, a training-free KV-cache eviction framework for long-horizon LLM agents. MemDecay assigns each token a region label supplied by the orchestration layer, combines region-specific base priorities with attention-derived importance, and applies region-specific temporal decay to each token's retention priority. Under a fixed cache budget, MemDecay evicts the pages holding the lowest-priority tokens while protecting pinned instruction regions. Region labels align with contiguous prompt spans, so the policy maps onto page-granular cache layouts in existing serving stacks.

We further describe a measurement protocol that estimates region-conditioned attention lifetimes from agent traces and uses them to calibrate decay rates. In a controlled study spanning two context lengths (roughly 450 and 1{,}700 tokens) and two model sizes (Qwen2.5-1.5B and 3B), measured lifetimes differ by an order of magnitude across regions in every setting (system half-lives of 148 to 189 decode steps against 14 to 16 for scratchpad reasoning), and the ordering of the movable regions is robust to insertion order. The eviction study yields one clear win and one instructive loss. Pinning preserves system-region facts at the full-cache ceiling in every setting, no baseline preserves more than 13 of 24, and recency-based retention collapses entirely as contexts grow while the structural prior keeps working. On unpinned content, accumulated-attention retention strengthens with model scale and wins; an ablation shows the policy's attention term cannot compensate at any tested weight, pointing to magnitude normalization of that term as the needed design change.
\end{abstract}




\setlength{\parskip}{0.7em}
\setlength{\parindent}{0pt}


\section{Introduction}

Large language models (LLMs) are increasingly used in long-horizon agentic workflows that combine multi-turn interaction, retrieval, tool use, and iterative planning. Unlike single-pass long-context tasks, these systems accumulate a persistent execution trace composed of system instructions, safety constraints, evolving plans, user turns, retrieved evidence, tool interactions, and intermediate reasoning states. The trace is structured: its components serve different functional roles and should persist for different durations.

This shift places new pressure on the key--value (KV) cache that supports autoregressive decoding. Because the KV cache grows with sequence length, long interaction traces translate into substantial memory overhead, increased bandwidth demand, and higher end-to-end latency \cite{vllm,kv_survey}. Prior work reduces this overhead through selective retention and compression, including policies that adapt eviction to stable differences in attention behavior across heads \cite{fastgen,adakv}. Most of these approaches target generic long-context generation and score cached tokens with model-centric signals such as accumulated attention, attention sinks and recency windows, or instantaneous attention \cite{h2o,streamingllm,tova}.

That assumption is poorly matched to agentic workloads. In LLM-based agents, memory is a core component that supports long-term and complex interactions with the environment \cite{agent_memory_survey}. Yet the contents of an agent trace are heterogeneous by design. System and safety instructions often need to remain stable throughout an episode, and high-level plans may need to persist until the agent replans. By contrast, stale tool outputs, transient scratchpad content, and low-value retrievals are often useful only for a short span of the trajectory. A cache policy that treats all such tokens as having the same expected lifetime misses this structure.

This mismatch motivates aligning cache retention with the semantic structure already present in orchestrated prompts. We propose \textbf{MemDecay}, a training-free, prompt-region-aware KV-cache decay and eviction framework for long-horizon LLM agents. MemDecay assumes that the orchestration layer provides region labels for tokens, such as system prompt, plan, user message, tool input, tool output, retrieval, or scratchpad. It assigns each region a base retention priority and a decay rate, then combines these structural priors with attention-derived importance to compute a composite retention score for each token. MemDecay is training-free in the sense this literature uses the term: it updates no model weights and learns no predictor. It does rely on a lightweight statistical calibration that fits two parameters per region to measured traces, described in Section~\ref*{sec:experimental}. Concurrent work gives evidence that semantic roles predict reuse and attention lifetimes in agent workloads \cite{saecache,rolekv}; MemDecay applies this signal to token-scored, page-granular eviction under a fixed budget within a single decoding context.

MemDecay treats forgetting as a scheduled process. Region-specific temporal decay lowers a token's retention priority smoothly over time, and attention to the token resets its decay clock, so continued use keeps a token alive. When a fixed KV-cache budget is exceeded, tokens with the lowest decayed priorities are evicted first, while pinned instruction regions and, optionally, active plan segments are preserved. The resulting policy is interpretable, requires no retraining or auxiliary modules, and operates on page-granular cache layouts of the kind used in existing serving stacks \cite{vllm}.

Our contributions are threefold:
\begin{itemize}
    \item We formulate region-aware selective forgetting as a KV-cache memory-management problem for long-horizon LLM agents, grounded in the structure of orchestrated prompts and agent traces.
    \item We introduce \textbf{MemDecay}, a training-free, prompt-region-aware eviction policy that combines region base priorities, region-specific temporal decay with attention-triggered refresh, and attention-derived importance in a single interpretable retention score under a fixed KV-cache budget.
    \item We specify a measurement protocol that estimates region-conditioned attention lifetimes from agent traces and uses the fitted decay rates to calibrate the policy, and we execute it across two context lengths and two model sizes (4{,}320 scored eviction probes): the region-lifetime hierarchy replicates in every setting, with the longest- and shortest-lived regions separated by an order of magnitude at non-overlapping confidence intervals, the fitted rates overturn one qualitative default, pinning delivers guaranteed instruction survival everywhere, recency-based retention collapses with context length while the structural prior keeps working, and an invariance ablation points to magnitude normalization of the attention term as the needed design change for the policy's unpinned-recall loss.
\end{itemize}

\section{Related Work}

\subsection{KV-Cache Memory Management for LLM Serving}

KV-cache management has become a first-order systems problem for scalable LLM serving because cache footprint grows with context length and directly constrains memory capacity, bandwidth, and throughput \cite{vllm,kv_survey}. System-level approaches focus on how cached states are allocated, retained, and reused across requests. PagedAttention organizes the cache into fixed-size pages to reduce fragmentation and enable sharing \cite{vllm}, and prefix-aware designs detect and reuse shared prompt segments such as system instructions and retrieved passages \cite{chunkattention,cacheblend}. Studies of production-style prefix caches report that reuse value varies widely across prompt segments, which motivates retention decisions that account for what a segment contains \cite{saecache}.

\subsection{KV-Cache Eviction and Compression for Long-Context LLMs}

A large literature studies post-training KV reduction through eviction, pruning, quantization, and hybrid schemes for long-context inference \cite{kv_survey,kv_review}. Attention-based eviction retains tokens with high accumulated attention \cite{h2o,scissorhands}, window-based policies keep attention sinks and recent tokens \cite{streamingllm,tova}, and prompt-time compression selects tokens using attention from an observation window \cite{snapkv}.

Later work showed that locally accumulated attention statistics are biased and can misestimate future utility \cite{nacl,expected_attention}. Several methods reallocate budgets across heads or layers, reflecting evidence that attention modules have heterogeneous roles and sensitivities \cite{fastgen,adakv}. Others score tokens with query-aware or semantic-cluster criteria \cite{quest,clusterkv}, keep or drop contiguous semantic chunks instead of isolated tokens \cite{chunkkv}, or lower the precision of cached entries instead of removing them \cite{kivi}. Learned approaches predict token importance directly, at the cost of training and auxiliary modules \cite{lkv,saecache}.

\subsection{Agent- and Workflow-Aware KV Management}

Recent work shows that agentic workloads differ materially from static chat or document settings because they interleave repeated LLM calls, tools, and variable pauses, creating reuse patterns that generic policies do not anticipate \cite{continuum}. Workflow-aware systems model future execution structure explicitly. KVFlow prioritizes retention using estimated temporal proximity to future agent activation, and Continuum uses time-to-live pinning to preserve KV states across short tool calls when recomputation would be costly \cite{kvflow,continuum}. Prefix-cache studies report that segment semantics matter in practice: token-agnostic policies such as LRU miss up to 756$\times$ variation in reuse rates across system prompts, user queries, tool outputs, model responses, and reasoning traces \cite{saecache}.

Concurrent work brings role and time structure into cache policies themselves. RoleKV tags KV blocks with semantic roles and applies role-calibrated decay to serving-layer retention, reporting that the oldest blocks in agent traces, including system prompts and tool outputs, receive the most persistent attention while recent reasoning tokens decay fastest \cite{rolekv}. IntentKV scores history tokens against an exponentially decayed summary of session queries for multi-turn agent pruning \cite{intentkv}, and retention-gated caching learns per-token gates that decay token importance over time \cite{trimkv}. These results establish that semantic roles and temporal decay both carry signal for cache management, and they disagree about which regions of an agent trace stay useful, a question we return to in Section~\ref*{sec:experimental}.

\subsection{Memory and Forgetting}

Memory is a defining component of LLM-based agents because long-horizon interaction depends on retaining, updating, and selectively reusing prior state \cite{agent_memory_survey,memory_survey}. Within KV-cache research, several recent papers argue that irreversible keep-or-drop policies are too coarse: token value can shift as decoding proceeds, and a token evicted early cannot be recovered when later queries make it relevant again \cite{intentkv,confkv}. Mixed-precision storage offers one response, degrading uncertain entries before removing them \cite{confkv,kivi}. This body of work motivates treating forgetting as a controllable mechanism with an explicit schedule instead of a byproduct of memory pressure. Class-based replacement with aging has a long history in web caching, where Greedy-Dual-Size-Frequency combines per-class value with an aging term \cite{gdsf}; MemDecay adapts this lineage to KV caches, with prompt regions as classes and decay rates calibrated per region.

\subsection{Positioning MemDecay}

MemDecay is closest to the concurrent role- and time-aware policies above, and it differs from each on a specific axis. RoleKV manages which blocks stay resident for cross-request reuse and is lossless with respect to model outputs \cite{rolekv}; MemDecay performs lossy, page-granular eviction driven by token-level scores inside a single decoding context under a fixed budget. SAECache learns segment reuse value online for prefix caches \cite{saecache}, and IntentKV in its stronger variant trains a residual scoring head \cite{intentkv}; MemDecay is training-free and uses only signals available at inference time. Retention-gated caching learns its decay behavior through trained per-token gates \cite{trimkv}; MemDecay is training-free, conditions both the base priority and the decay rate on the prompt region, and lets attention reset the decay clock. Workflow-level systems such as KVFlow and Continuum schedule caches across agent steps and are complementary: MemDecay decides which tokens survive within a context, and those systems decide when whole caches are kept, offloaded, or dropped \cite{kvflow,continuum}.

\section{Proposed Method}

\begin{figure}[t]
\centering
\begin{tikzpicture}[
  font=\footnotesize,
  box/.style={draw, rounded corners, align=center, inner sep=4pt},
  seg/.style={draw, minimum height=13pt, inner sep=3pt, font=\scriptsize\ttfamily},
  arr/.style={-{Stealth}, semithick}
]
\node[box] (orch) {Agent\\orchestrator};
\node[box, right=1.7cm of orch] (llm) {LLM\\decode};
\node[seg, fill=black!25, below=1.0cm of orch] (c1) {sys};
\node[seg, fill=black!15, right=0cm of c1] (c2) {plan};
\node[seg, fill=black!10, right=0cm of c2] (c3) {user};
\node[seg, right=0cm of c3] (c4) {tool};
\node[seg, right=0cm of c4] (c5) {retr};
\node[seg, right=0cm of c5] (c6) {scr};
\node[right=2pt of c6, font=\scriptsize, align=left] {KV cache\\(pages)};
\node[box, below=1.0cm of c3.south east, text width=0.8\columnwidth] (md)
  {\textbf{MemDecay:} attention EWMA (\ref*{eq:ewma}), decay clocks (\ref*{eq:refresh}), retention scores (\ref*{eq:score})};
\draw[arr] (orch) -- node[above, font=\scriptsize, align=center]{prompt +\\region labels} (llm);
\draw[arr] (llm.south) -- node[right, font=\scriptsize]{append KV} (llm.south |- c6.north);
\draw[arr] (c2.south) -- node[left, font=\scriptsize, align=center]{attention\\statistics} (c2.south |- md.north);
\draw[arr] (md.north -| c5.south) -- node[right, font=\scriptsize, align=center]{evict lowest-\\score pages} (c5.south);
\end{tikzpicture}
\caption{MemDecay in the serving stack. The orchestrator labels each prompt segment with its region, the decode engine appends KV entries to region-aligned pages, and MemDecay maintains attention statistics and decay clocks, evicting the lowest-scoring non-pinned pages when the budget binds.}
\label{fig:arch}
\end{figure}
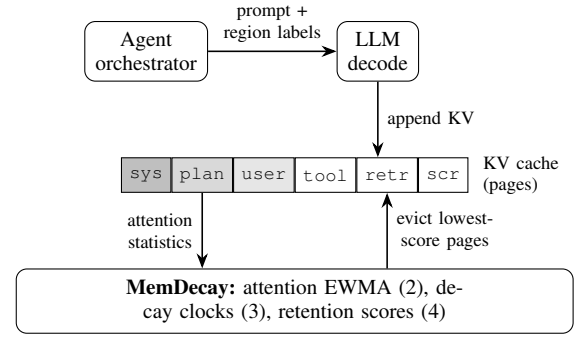

MemDecay is an inference-time eviction policy for the KV cache of a decoder-only LLM running inside an agent orchestrator. The policy uses three inputs: region labels supplied by the orchestrator, attention statistics sampled during decoding, and a fixed cache budget. Fig.~\ref*{fig:arch} shows where the policy sits in the serving stack. This section defines the setting, the structural priors, the retention score, and the eviction procedure.

\subsection{Problem Setup}
\label{sec:problem}

An agent episode produces a token sequence indexed by $i$, decoded over steps $t = 1, 2, \dots$. Each cached token $i$ carries a region label $r(i) \in \mathcal{R}$ and an insertion step $t_i^{\mathrm{ins}}$, where $\mathcal{R}$ contains the labels \texttt{system}, \texttt{plan}, \texttt{user}, \texttt{tool\_in}, \texttt{tool\_out}, \texttt{retrieval}, and \texttt{scratchpad}. Region labels correspond to contiguous spans of the prompt because the orchestrator assembles the context from template segments. We assume the orchestration layer emits these labels alongside the request; serving systems have begun to accept exactly this kind of metadata, and an open SGLang proposal attaches workflow, tool, and reuse hints to inference requests \cite{sglang_rfc}.

Let $\mathcal{K}_t$ denote the set of cached tokens at step $t$ and let $B$ be the cache budget in tokens. A subset $\mathcal{F} \subseteq \mathcal{K}_t$ of pinned tokens is never evicted; by default $\mathcal{F}$ contains all \texttt{system} tokens, which also covers the initial attention-sink positions that must remain resident for stable decoding \cite{streamingllm}. The policy must maintain $|\mathcal{K}_t| \leq B$ by selecting which non-pinned tokens to evict. Formally, at any step where the budget binds, the policy solves
\begin{equation}
\max_{\mathcal{K} \subseteq \mathcal{K}_t} \; \sum_{i \in \mathcal{K}} s_i(t)
\quad \text{s.t.} \quad |\mathcal{K}| \leq B, \;\; \mathcal{F} \subseteq \mathcal{K},
\label{eq:objective}
\end{equation}
where $s_i(t)$ is the retention score defined in Section~\ref*{sec:score}. The objective is additive over tokens, so retaining the $B$ highest-scoring tokens, after forcing $\mathcal{F}$ into the set, is optimal, and eviction reduces to removing the lowest-scoring tokens first.

\subsection{Structural Priors}

Each region $r$ is configured with a base priority $b_r \geq 0$ and a decay rate $\lambda_r \geq 0$. The base priority encodes how valuable a token from region $r$ is at insertion time, and the decay rate encodes how quickly that structural protection expires. Table~\ref*{tab:regions} lists the default configuration. These defaults encode the template lifetimes described in Section~I. They are starting points: published evidence on region lifetimes conflicts, with role-aware serving measurements reporting persistent attention to old system prompts and tool outputs \cite{rolekv} while decay-based retention assumes that older tokens generally matter less \cite{trimkv}. Section~\ref*{sec:experimental} describes a calibration procedure that replaces these defaults with rates fitted per workload.

\begin{table}[t]
\caption{Default region configuration. Calibration (Section~\ref*{sec:experimental}) replaces the qualitative decay rates with per-workload fitted values.}
\label{tab:regions}
\centering
\small
\begin{tabular}{lccc}
\hline
Region & Base priority $b_r$ & Decay rate $\lambda_r$ & Pinned \\
\hline
\texttt{system} & high & $\approx 0$ & yes \\
\texttt{plan} & high & small & optional \\
\texttt{user} & medium & medium & no \\
\texttt{tool\_in} & low & large & no \\
\texttt{tool\_out} & medium & calibrated & no \\
\texttt{retrieval} & low & large & no \\
\texttt{scratchpad} & low & large & no \\
\hline
\end{tabular}
\end{table}

\subsection{Retention Score with Region-Specific Decay}
\label{sec:score}

MemDecay tracks one dynamic signal per token, an attention importance $a_i$, and one clock per token, the time since the token was last used. Fused attention kernels do not materialize the attention matrix, so reading per-step attention scores is impractical in production kernels \cite{nvidia_kv_blog}. MemDecay therefore samples attention only at observation steps, one every $k$ decode steps, using windowed scores in the style of prompt-time selection methods \cite{snapkv}. At an observation step, let $\bar{A}_i$ denote the attention mass received by token $i$, averaged over heads, layers, and the window. The importance is an exponentially weighted average,
\begin{equation}
a_i \leftarrow \rho\, a_i + (1 - \rho)\, \bar{A}_i,
\label{eq:ewma}
\end{equation}
with smoothing factor $\rho \in [0, 1)$.

The decay clock measures time since insertion or since the model last attended to the token above a threshold $\tau$. Let
\begin{equation}
\Delta t_i(t) = t - t_i^{\mathrm{ref}}, \qquad
t_i^{\mathrm{ref}} =
\begin{cases}
t' & \text{if } \bar{A}_i \geq \tau \text{ at step } t', \\
t_i^{\mathrm{ins}} & \text{otherwise},
\end{cases}
\label{eq:refresh}
\end{equation}
where $t'$ is the most recent observation step satisfying the condition. A refresh resets the clock and nothing else; the structural prior is re-applied from the refresh point. The threshold $\tau$ can be fixed or scaled to the uniform attention level $1/|\mathcal{K}_t|$ so that the refresh criterion tracks context length; Section~\ref*{sec:experimental} uses the scaled form.

\begin{figure}[t]
\centering
\begin{tikzpicture}
\begin{axis}[
  width=\columnwidth, height=0.75\columnwidth,
  xlabel={token age $\Delta t_i$ (decode steps)},
  ylabel={retention score $s_i(t)$},
  xmin=0, xmax=100, ymin=0, ymax=1.05,
  legend style={font=\scriptsize, at={(0.5,-0.42)}, anchor=north, draw=none, fill=none, /tikz/every even column/.append style={column sep=0.35cm}},
  legend columns=2,
  tick label style={font=\scriptsize},
  label style={font=\footnotesize},
  domain=0:100, samples=120, no markers,
  every axis plot/.append style={thick, black}
]
\addplot[solid] {1.0*exp(-0.002*x)};
\addlegendentry{\texttt{system}: $b{=}1.0$, $\lambda{=}0.002$}
\addplot[dashed] {0.9*exp(-0.01*x)};
\addlegendentry{\texttt{plan}: $b{=}0.9$, $\lambda{=}0.01$}
\addplot[dotted] {0.6*exp(-0.03*x)};
\addlegendentry{\texttt{user}: $b{=}0.6$, $\lambda{=}0.03$}
\addplot[dashdotted] {0.4*exp(-0.08*x)};
\addlegendentry{\texttt{retrieval}: $b{=}0.4$, $\lambda{=}0.08$}
\addplot[densely dashed, gray, domain=0:40] {0.6*exp(-0.08*x)};
\addlegendentry{\texttt{tool\_out}: refresh at $\Delta t{=}40$}
\addplot[densely dashed, gray, domain=40:100, forget plot] {0.6*exp(-0.08*(x-40))};
\end{axis}
\end{tikzpicture}
\caption{Retention score trajectories from (\ref*{eq:score}) under example parameters, with $\alpha a_i = 0$ for clarity; the curves illustrate the equation and are not measurements. Pinned \texttt{system} tokens decay negligibly and \texttt{plan} tokens decay slowly, while short-lived regions decay quickly unless attention above the threshold $\tau$ resets the decay clock, shown for \texttt{tool\_out} at age $40$.}
\label{fig:decay}
\end{figure}
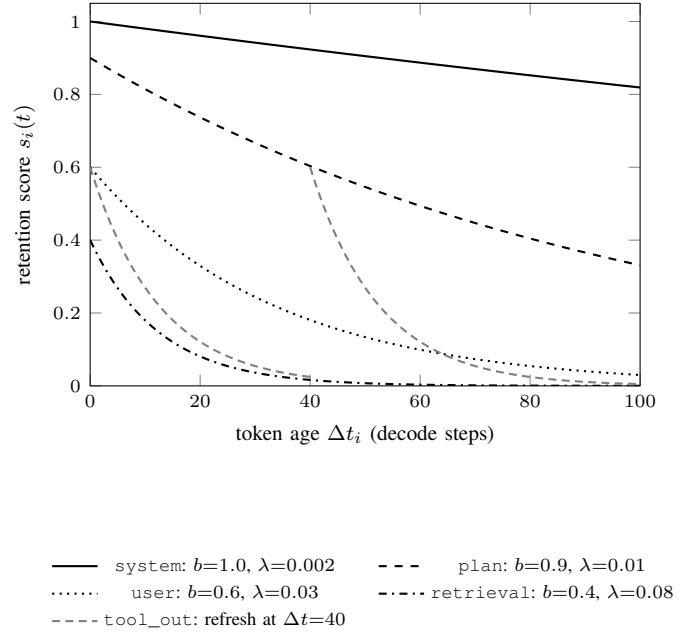

The retention score combines the decayed structural prior with the observed importance:
\begin{equation}
s_i(t) = b_{r(i)} \exp\!\left(-\lambda_{r(i)}\, \Delta t_i(t)\right) + \alpha\, a_i(t),
\label{eq:score}
\end{equation}
where $\alpha \geq 0$ weights how much observed attention can protect a token beyond its structural prior. Recency enters only through the decay clock, so the score contains no separate recency term and elapsed time is not counted twice. The three terms have direct interpretations: $b_r$ sets where a token starts, $\lambda_r$ sets how quickly structure alone stops protecting it, and $\alpha a_i$ keeps any token alive while the model keeps using it, regardless of region. Pinned tokens bypass scoring entirely. Fig.~\ref*{fig:decay} illustrates the resulting score trajectories, including a refresh event that resets the decay clock of a short-lived region.

\subsection{Budget-Aware Eviction at Page Granularity}

Production serving stacks store KV entries in fixed-size pages \cite{vllm}. Region labels mark contiguous spans, so at most one page can straddle each boundary between consecutive region runs, and the straddle fraction shrinks as regions grow long relative to the page size. In the episodes of Section~\ref*{sec:experimental}, 35\% of 16-token pages straddle a boundary on the short tier and 17\% on the long tier, matching the prediction that straddling shrinks as regions lengthen. The page score is the mean over member tokens,
\begin{equation}
S_p(t) = \frac{1}{|p|} \sum_{i \in p} s_i(t).
\label{eq:page}
\end{equation}
MemDecay evicts at page granularity: when $|\mathcal{K}_t| > B$, non-pinned pages are ordered by ascending $S_p(t)$ and evicted until the budget is met. This greedy rule is optimal for the page-level objective because pages have equal size; it can deviate from the token-level optimum of (\ref*{eq:objective}) when a page mixes regions, since the mean in (\ref*{eq:page}) can mask a high-scoring token among low-scoring pagemates. The straddle fraction above quantifies how often mixed pages occur, and the pinned-page collateral disclosed in Section~\ref*{sec:setup} is the same effect acting in the protective direction. Region-aligned page boundaries, where the allocator opens a fresh page at each region boundary, would remove the dilution at the cost of internal fragmentation; we leave that variant to future work. Algorithm~\ref*{alg:memdecay} summarizes one observation step. Per-token state is two scalars ($a_i$ and $t_i^{\mathrm{ref}}$), updates occur only at observation steps, and eviction uses a min-heap over pages, so the policy adds $O(P \log P)$ work for $P$ pages at each observation step.

\begin{algorithm}[t]
\caption{MemDecay update and eviction at an observation step $t$}
\label{alg:memdecay}
\begin{algorithmic}[1]
\State Update $a_i$ for all cached tokens using (\ref*{eq:ewma})
\State Update refresh clocks $t_i^{\mathrm{ref}}$ using (\ref*{eq:refresh})
\State Compute $s_i(t)$ for all non-pinned tokens using (\ref*{eq:score})
\State Compute page scores $S_p(t)$ using (\ref*{eq:page})
\If{$|\mathcal{K}_t| > B$}
    \State Build a min-heap over non-pinned pages keyed by $S_p$
    \While{$|\mathcal{K}_t| > B$}
        \State Evict the page with minimum $S_p$
    \EndWhile
\EndIf
\end{algorithmic}
\end{algorithm}

\subsection{Scope and Tiered Extension}
\label{sec:tiered}

The decay in (\ref*{eq:score}) acts on retention priority. Retained tokens participate in attention with unchanged weights, and eviction remains a discrete, irreversible step in the base policy. The decay schedule therefore controls the order in which tokens become eviction candidates; it does not scale their influence on the model output. A tiered extension would provide a graded degradation path: pages whose score falls below a first threshold are quantized to low precision \cite{kivi}, pages below a second threshold are offloaded to host memory with background prefetch \cite{kvflow}, and pages below a final threshold are dropped. Mixed-precision retention keyed to token-level confidence has precedent in long-horizon settings \cite{confkv}. The tiered variant would recover offloaded pages if later queries make them relevant again, at the cost of host memory and transfer bandwidth.

\section{Evaluation}
\label{sec:experimental}

This section reports a controlled study across two context lengths and two model sizes, executed end to end by the released harness: every number below comes from the raw result files at \texttt{github.com/venkateshamatam/memdecay}, and the study totals 4{,}320 scored eviction probes plus three full measurement passes. The findings are stated the way the evidence supports them. The region-lifetime measurement replicates at every scale and is the primary result. The eviction policy has one clear win, guaranteed instruction survival through pinning, and one clear loss, unpinned recall against attention-based retention, and the loss has a diagnosed mechanism whose fix the ablation narrows down.

\subsection{Setup}
\label{sec:setup}

Episodes come in two tiers built from the same eight scenarios, each in three value variants with identical planted facts and probes. The short tier caches 405 to 507 tokens per episode (five stages); the long tier caches 1{,}656 to 1{,}723 tokens (eight stages, adding a second retrieved document, a third tool exchange, and a further user turn). In both tiers the movable blocks (tool exchanges, follow-up turns, retrieved documents) arrive in a different order in each variant, so region identity is decoupled from insertion position for the movable regions; the opening stage (system, plan, user) and the closing scratchpad seed keep fixed positions, which matches how orchestrators assemble contexts. Three settings are evaluated: Qwen2.5-1.5B-Instruct on the short tier, the same model on the long tier (both float32 on an Apple M3 Pro), and Qwen2.5-3B-Instruct on the long tier (float16 on a T4 GPU). All runs use eager attention, which is required to read attention weights \cite{nvidia_kv_blog}. A numerical audit of the measurement pipeline bounded the effect of half precision on fitted decay rates below 0.001\% before the 3B run, and the measurement pipeline is deterministic across devices: the long tier's per-episode straddle statistics reproduce exactly between the M3 Pro and the T4.

Each episode plants four facts with arbitrary values in the \texttt{system}, \texttt{user}, \texttt{tool\_out}, and \texttt{retrieval} regions and pairs each with a probe question, answered by greedy decoding of 16 tokens after eviction. Recall requires the gold value to appear in the generated text at token boundaries, and each configuration cell aggregates 96 probes (24 episodes, 4 probes). The probes arrive after compression, so the setting is query-agnostic, which prior work identifies as substantially harder than compressing with the query in hand \cite{kvzip}. Statistics cluster at the scenario level: variants of one scenario are not independent, so confidence intervals come from a percentile bootstrap over the eight scenarios and are approximate, and the paired per-scenario sign comparison is the statistic the policy claims rest on. The random baseline runs three seeds, collapsed to one value per probe before statistics.

MemDecay uses the Table~\ref*{tab:regions} base priorities, $\rho = 0.9$, $\alpha = 0.5$, $\tau = 2/|\mathcal{K}_t|$, page size 16, and a pinned \texttt{system} region; pinning protects whole pages, so tokens sharing a page with \texttt{system} tokens are protected as well. Decay rates are calibrated leave-one-scenario-out within each setting, so no sibling variant leaks into its own calibration. Two calibration details matter for interpretation. The decay rates are fitted under a 48-step final observation burst, while the signals available at eviction time come from an 8-step burst; this asymmetry is identical across all policies and tiers. The count-weighted fit also anchors its intercept toward the observation-dense low-age buckets, again constant across conditions. Attention is sampled at every decode step ($k = 1$); larger observation intervals are the production setting. Budgets are fractions of each episode's cache, so the 25\% condition sets $B = \lfloor 0.25\,n \rfloor$.

\subsection{Measured Region Lifetimes Across Scales}
\label{sec:lifetimes}

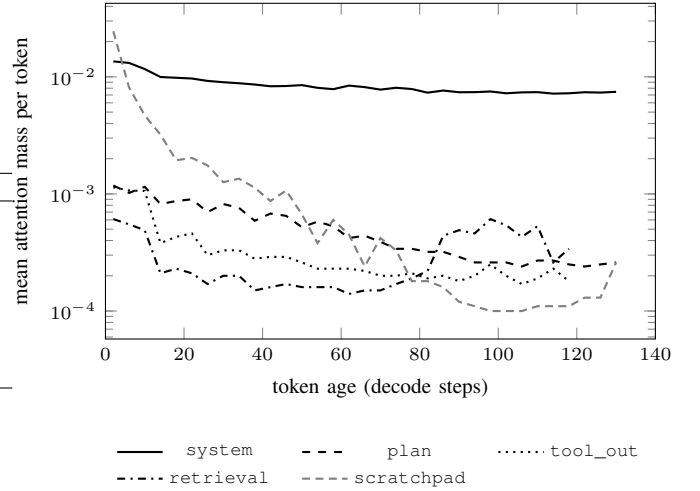
\begin{figure}[t]
\centering
\begin{tikzpicture}
\begin{axis}[
  width=\columnwidth, height=0.68\columnwidth,
  xlabel={token age (decode steps)},
  ylabel={mean attention mass per token},
  ymode=log,
  xmin=0, xmax=140,
  legend style={font=\scriptsize, at={(0.5,-0.28)}, anchor=north, draw=none, fill=none, /tikz/every even column/.append style={column sep=0.35cm}},
  legend columns=3,
  tick label style={font=\scriptsize},
  label style={font=\footnotesize},
  no markers, every axis plot/.append style={thick, black}
]
\addplot[solid] coordinates {(2,0.01355) (6,0.01315) (10,0.01166) (14,0.00997) (18,0.00982) (22,0.00968) (26,0.00923) (30,0.00900) (34,0.00882) (38,0.00860) (42,0.00831) (46,0.00835) (50,0.00850) (54,0.00806) (58,0.00785) (62,0.00842) (66,0.00817) (70,0.00778) (74,0.00807) (78,0.00787) (82,0.00733) (86,0.00764) (90,0.00739) (94,0.00740) (98,0.00750) (102,0.00724) (106,0.00737) (110,0.00740) (114,0.00719) (118,0.00724) (122,0.00739) (126,0.00734) (130,0.00745)};
\addlegendentry{\texttt{system}}
\addplot[dashed] coordinates {(2,0.00118) (6,0.00102) (10,0.00115) (14,0.00082) (18,0.00087) (22,0.00090) (26,0.00070) (30,0.00082) (34,0.00076) (38,0.00059) (42,0.00068) (46,0.00065) (50,0.00052) (54,0.00058) (58,0.00053) (62,0.00042) (66,0.00044) (70,0.00039) (74,0.00034) (78,0.00034) (82,0.00032) (86,0.00032) (90,0.00029) (94,0.00026) (98,0.00026) (102,0.00026) (106,0.00024) (110,0.00027) (114,0.00027) (118,0.00025) (122,0.00024) (126,0.00025) (130,0.00026)};
\addlegendentry{\texttt{plan}}
\addplot[dotted] coordinates {(2,0.00114) (6,0.00107) (10,0.00106) (14,0.00038) (18,0.00043) (22,0.00046) (26,0.00030) (30,0.00033) (34,0.00033) (38,0.00028) (42,0.00029) (46,0.00029) (50,0.00026) (54,0.00023) (58,0.00023) (62,0.00023) (66,0.00022) (70,0.00020) (74,0.00020) (78,0.00021) (82,0.00019) (86,0.00020) (90,0.00018) (94,0.00020) (98,0.00025) (102,0.00020) (106,0.00017) (110,0.00019) (114,0.00023) (118,0.00018)};
\addlegendentry{\texttt{tool\_out}}
\addplot[dashdotted] coordinates {(2,0.00061) (6,0.00055) (10,0.00049) (14,0.00021) (18,0.00023) (22,0.00021) (26,0.00017) (30,0.00020) (34,0.00020) (38,0.00015) (42,0.00016) (46,0.00017) (50,0.00016) (54,0.00016) (58,0.00016) (62,0.00014) (66,0.00015) (70,0.00015) (74,0.00017) (78,0.00019) (82,0.00022) (86,0.00044) (90,0.00049) (94,0.00046) (98,0.00061) (102,0.00054) (106,0.00043) (110,0.00053) (114,0.00026) (118,0.00034)};
\addlegendentry{\texttt{retrieval}}
\addplot[densely dashed, gray] coordinates {(2,0.02450) (6,0.00813) (10,0.00469) (14,0.00321) (18,0.00194) (22,0.00203) (26,0.00175) (30,0.00126) (34,0.00135) (38,0.00113) (42,0.00087) (46,0.00107) (50,0.00067) (54,0.00038) (58,0.00060) (62,0.00045) (66,0.00024) (70,0.00042) (74,0.00033) (78,0.00018) (82,0.00018) (86,0.00016) (90,0.00012) (94,0.00011) (98,0.00010) (102,0.00010) (106,0.00010) (110,0.00011) (114,0.00011) (118,0.00011) (122,0.00013) (126,0.00013) (130,0.00026)};
\addlegendentry{\texttt{scratchpad}}
\end{axis}
\end{tikzpicture}
\caption{Measured mean attention mass per token by region and token age for the largest setting (Qwen2.5-3B, long tier, 24 episodes; age buckets of 4 steps, log scale; point estimates). \texttt{system} stays persistently attended, \texttt{scratchpad} decays fastest, and \texttt{retrieval} outlives every other movable region at each of its three insertion positions. Table~\ref*{tab:fits} shows the same hierarchy in the other two settings.}
\label{fig:measured}
\end{figure}

Let $\bar{A}_r(\Delta t)$ denote the mean attention mass received by tokens of region $r$ at age $\Delta t$, aggregated over episodes. The calibration fits the log-linear model
\begin{equation}
\log \bar{A}_r(\Delta t) \approx \log c_r - \hat{\lambda}_r \, \Delta t
\label{eq:fit}
\end{equation}
by weighted least squares and sets $\lambda_r \leftarrow \hat{\lambda}_r$; the associated half-life
\begin{equation}
t_r^{1/2} = \ln 2 / \hat{\lambda}_r
\label{eq:halflife}
\end{equation}
summarizes how long region $r$ remains useful. Table~\ref*{tab:fits} reports the fitted half-lives with scenario-bootstrap confidence intervals for all three settings, and Fig.~\ref*{fig:measured} shows the measured curves for the largest one.

\begin{table}[t]
\caption{Fitted attention half-lives in decode steps (Eqs.~\ref*{eq:fit}--\ref*{eq:halflife}), with 95\% scenario-bootstrap confidence intervals, across the three settings. $R^2$ spans 0.30 to 0.95; it is lowest for \texttt{retrieval}, whose near-flat curves leave little age-driven variance to explain.}
\label{tab:fits}
\centering
\small
\begin{tabular}{lccc}
\hline
Region & 1.5B short & 1.5B long & 3B long \\
\hline
\texttt{system} & 150 [126, 182] & 148 [136, 166] & 189 [174, 206] \\
\texttt{plan} & 33 [27, 38] & 44 [40, 49] & 52 [49, 55] \\
\texttt{user} & 27 [24, 31] & 31 [27, 35] & 35 [30, 40] \\
\texttt{tool\_in} & 26 [21, 32] & 22 [19, 25] & 26 [22, 29] \\
\texttt{tool\_out} & 33 [27, 38] & 37 [32, 42] & 39 [32, 50] \\
\texttt{retrieval} & 48 [38, 61] & 76 [59, 95] & 66 [50, 92] \\
\texttt{scratchpad} & 14 [13, 15] & 14 [13, 15] & 16 [15, 17] \\
\hline
\end{tabular}
\end{table}

The region hierarchy is stable everywhere it was measured. \texttt{system} outlives \texttt{scratchpad} by an order of magnitude in every setting, with non-overlapping intervals, and becomes more persistent on the larger model (189 against 148 at matched context). \texttt{scratchpad} churns at a half-life of 14 to 16 steps regardless of model or context length. \texttt{retrieval} is the longest-lived movable region in every setting, even though it arrives at three different positions across a scenario's variants, and it holds attention longer at the longer context; this again contradicts the fast-decay default that Table~\ref*{tab:regions} assigns it, which is the calibration argument made concrete. The straddle statistic also behaved as Section~III-D's bound predicts: 35\% of 16-token pages straddle a region boundary on the short tier and 17\% on the long tier, since longer regions span more page-interior tokens.

\subsection{Budget-Constrained Recall Across Scales}
\label{sec:recall}

Table~\ref*{tab:recall} reports the short-tier recall and Table~\ref*{tab:recall_long} the long-tier recall for both models, after eviction to 25\% and 50\% of the cache, overall and excluding the 24 probes whose answers live in the pinned \texttt{system} region. Baselines are the unconstrained full cache, \emph{random} (4 sinks, 16 most recent, uniform sample, three seeds), \emph{streaming} (4 sinks plus the most recent tokens \cite{streamingllm}), and \emph{h2o}, a reimplementation of accumulated-attention retention \cite{h2o} restricted to the same observation signal as MemDecay. Baselines receive no region labels, and the reimplementations are not claims about the original methods at scale. MemDecay evicts whole pages, so it retained slightly fewer tokens than every baseline in all settings (for example 24.5\% versus 25.0\% on the long tier).

\begin{table}[t]
\caption{Short-tier recall (Qwen2.5-1.5B, 405--507-token episodes) after eviction, with 95\% cluster-bootstrap confidence intervals over 8 scenarios (96 probes per cell; random averages 3 seeds). ``Non-\texttt{sys}'' excludes probes answered in the pinned \texttt{system} region.}
\label{tab:recall}
\centering
\small
\begin{tabular}{llcc}
\hline
Policy & Budget & All & Non-\texttt{sys} \\
\hline
Full cache & 100\% & 0.97 [0.91, 1.00] & 0.96 [0.88, 1.00] \\
\hline
Random & 25\% & 0.20 [0.14, 0.26] & \textbf{0.23 [0.16, 0.29]} \\
Streaming & 25\% & 0.16 [0.08, 0.23] & 0.21 [0.11, 0.31] \\
H2O-style & 25\% & 0.15 [0.08, 0.21] & 0.19 [0.11, 0.28] \\
MemDecay & 25\% & \textbf{0.32 [0.28, 0.36]} & 0.10 [0.04, 0.15] \\
\hline
Random & 50\% & 0.38 [0.33, 0.43] & 0.37 [0.29, 0.44] \\
Streaming & 50\% & 0.27 [0.21, 0.32] & 0.36 [0.28, 0.43] \\
H2O-style & 50\% & 0.32 [0.20, 0.45] & \textbf{0.40 [0.24, 0.57]} \\
MemDecay & 50\% & \textbf{0.49 [0.41, 0.55]} & 0.32 [0.21, 0.40] \\
\hline
\end{tabular}
\end{table}

\begin{table*}[t]
\caption{Long-tier recall (1{,}656--1{,}723-token episodes) after eviction, for both models, with 95\% cluster-bootstrap confidence intervals over 8 scenarios (96 probes per cell; random averages 3 seeds).}
\label{tab:recall_long}
\centering
\small
\begin{tabular}{llcccc}
\hline
 & & \multicolumn{2}{c}{Qwen2.5-1.5B} & \multicolumn{2}{c}{Qwen2.5-3B} \\
Policy & Budget & All & Non-\texttt{sys} & All & Non-\texttt{sys} \\
\hline
Full cache & 100\% & 0.91 [0.80, 1.00] & 0.93 [0.85, 1.00] & 0.91 [0.84, 0.97] & 0.92 [0.83, 0.99] \\
\hline
Random & 25\% & 0.17 [0.10, 0.24] & 0.19 [0.11, 0.25] & 0.12 [0.08, 0.18] & 0.10 [0.06, 0.14] \\
Streaming & 25\% & 0.10 [0.05, 0.16] & 0.11 [0.04, 0.18] & 0.01 [0.00, 0.03] & 0.01 [0.00, 0.04] \\
H2O-style & 25\% & 0.33 [0.20, 0.45] & \textbf{0.38 [0.22, 0.50]} & \textbf{0.39 [0.27, 0.49]} & \textbf{0.33 [0.21, 0.44]} \\
MemDecay & 25\% & \textbf{0.38 [0.31, 0.44]} & 0.18 [0.10, 0.25] & 0.32 [0.27, 0.39] & 0.12 [0.03, 0.26] \\
\hline
Random & 50\% & 0.31 [0.26, 0.36] & 0.35 [0.29, 0.41] & 0.27 [0.24, 0.31] & 0.25 [0.20, 0.31] \\
Streaming & 50\% & 0.10 [0.05, 0.16] & 0.12 [0.06, 0.21] & 0.03 [0.01, 0.06] & 0.03 [0.00, 0.07] \\
H2O-style & 50\% & \textbf{0.46 [0.28, 0.61]} & \textbf{0.53 [0.33, 0.71]} & \textbf{0.58 [0.45, 0.72]} & \textbf{0.60 [0.47, 0.72]} \\
MemDecay & 50\% & 0.43 [0.32, 0.53] & 0.28 [0.14, 0.42] & 0.39 [0.31, 0.46] & 0.22 [0.08, 0.40] \\
\hline
\end{tabular}
\end{table*}

Four findings organize the cross-scale picture.

First, the pinning guarantee holds everywhere. MemDecay answers 24 of 24 system probes on the short tier and 21 of 24 on the long tier at the half budget, at the full-cache ceiling (the models themselves miss some system probes at long context with nothing evicted), while streaming answers 0 to 2 of 24 in every setting. Guaranteed instruction survival under memory pressure is a designed property that no attention- or recency-based baseline provides.

Second, recency collapses with context length. Streaming is competitive on the short tier (0.36 non-system at the half budget, within the intervals of random at 0.37 and h2o at 0.40) and near zero on the long tier at both models (0.01 to 0.13), because its window physically cannot span the trace. MemDecay's paired per-scenario comparison against streaming flips accordingly: 0 of 8 scenarios positive on the short tier, then $+4/0$, $+5/1$, $+3/0$, and $+6/0$ across the four long-tier cells (mean deltas $+0.07$ to $+0.19$). Structural priors age gracefully where recency does not.

Third, accumulated attention strengthens with model scale. The h2o baseline is the weakest baseline on the short tier at the quarter budget, overtakes everything unpinned on the long tier at 1.5B, and dominates unpinned recall at 3B (0.60 non-system at the half budget, and 20 of 24 user probes), an external pattern consistent with published reports of H2O's task-dependence, which range from collapse on retrieval tasks \cite{dynamickv} to strength when its signal is informative.

Fourth, the mechanism behind MemDecay's unpinned loss is the same at every scale: the oldest unpinned facts score lowest under a decay prior. The user-region fact sits in the opening turn, and at the half budget MemDecay recalls 0 of 24 (short), 7 of 24 (1.5B long), and 5 of 24 (3B long) of them while h2o recalls 11 to 20 of 24; attention identifies exactly the tokens that structure ranks for eviction. The ablation below shows that no tested setting of the attention weight recovers this.

\subsection{Sensitivity and Overhead}
\label{sec:ablation}

An ablation on the short tier varies the attention weight $\alpha \in \{0.2, 0.5, 0.8\}$ and disables pinning, holding everything else fixed. The outcome is an invariance: all four configurations produce near-identical keep sets and near-identical probe outcomes (4 of 576 rows differ in kept count and one in recall, a single spurious flip that does not touch the user-region loss). Two scale facts explain it. The page mean divides any single token's attention contribution by the page size, and the EWMA with $\rho = 0.9$ has forgotten early attention by eviction time, so the $\alpha\,a_i$ term is one to two orders of magnitude below the structural term for exactly the old tokens that need rescue. Pinning also never binds: the calibrated \texttt{system} prior alone keeps its pages resident. The practical conclusion is that combining structural priors with attention requires magnitude normalization of the attention term (for example, relative to the uniform attention level), a design change rather than a tuning change; this is the concrete direction the recall results motivate.

Overhead was measured directly. The retention machinery itself is cheap: scoring plus page selection takes 0.27 ms and applying an eviction 4.2 ms on a 466-token cache. The cost is reading attention at all: with $k = 1$, a scored decode step takes 69.8 ms against 26.0 ms unscored on the M3 Pro (median decode steps during full measurement: 82.7 ms at 1.5B on the long tier on M3 Pro, 52.2 ms at 3B on a T4). Production deployments would sample attention at intervals ($k > 1$) or reuse scores produced for other purposes, and the eager-attention requirement this inherits from reading attention weights is shared with all attention-score methods \cite{nvidia_kv_blog}.

\section{Discussion and Limitations}

\textbf{Irreversibility.} The base policy cannot recover an evicted token when later context makes it relevant again, and such saliency shifts are documented in multi-turn agent traces \cite{intentkv}; the user-region result in Section~\ref*{sec:recall}, where the oldest unpinned facts are evicted and later probed, is a measured instance. The tiered extension of Section~\ref*{sec:tiered} would reduce this exposure by offloading before dropping.

\textbf{Cross-region information flow.} Agents copy content across regions, for example by quoting a tool output inside scratchpad reasoning. Under the default configuration both copies decay quickly, so the information can vanish from the cache even though the model produced and consumed it twice. Per-workload calibration reduces this risk; a mechanism that links copies to their sources is future work.

\textbf{Label quality.} MemDecay assumes region labels from the orchestrator. The serving-side metadata channel that would carry such labels is under active development \cite{sglang_rfc}, but workloads without a cooperating orchestrator would need a classifier to assign labels, and robustness to noisy labels is unmeasured.

\textbf{Granularity.} MemDecay applies one score per token, uniform across heads and layers. Head-adaptive budget allocation is orthogonal and could be composed with region-aware scoring \cite{fastgen,adakv}.

\textbf{Workload dependence.} Region lifetimes are properties of workloads, and published evidence disagrees across settings \cite{rolekv,trimkv}. Our own measurement reproduces this in miniature: the slow retrieval decay in Section~\ref*{sec:lifetimes} overturns the fast-decay default for that region. MemDecay treats the region configuration as measurable, which requires calibration data representative of the deployment workload.

\textbf{Scale and validity.} The study covers 1.5B and 3B models at roughly 450 and 1{,}700-token contexts, on heterogeneous hardware (Apple M3 Pro in float32, T4 GPU in float16, with the precision effect on fitted rates verified below 0.001\% and the measurement statistics reproducing exactly across devices). The 96 probes per cell derive from eight independent scenarios, so all intervals are cluster bootstraps over eight clusters and are approximate; the paired per-scenario sign comparison carries the policy claims. Interleaved schedules control insertion order for the movable regions, while system, plan, and scratchpad keep fixed positions, so their lifetime estimates remain position-confounded by construction. Contexts of 1{,}700 tokens still sit well below production long-context regimes, and conclusions about those regimes remain conditional on larger runs.

\textbf{Future work.} Three directions follow directly from the results. First, from static calibration to prediction: the ablation shows no weight on the current EWMA rescues old facts, so the attention term needs magnitude normalization (for example, relative to the uniform attention level) and a slower usage memory; the same machinery extends to decay rates that adapt online from observed refresh frequency per region, and to a lightweight future-reuse score built from age, region, phase, and attention trend, all still training-free. Second, scale: 8B-class models over contexts of 4k tokens and beyond, on real agent workloads (multi-turn coding and function-calling traces of the SWE-bench and BFCL kind \cite{continuum}, browsing traces \cite{intentkv}, and LongBench regression against compression baselines \cite{chunkkv}), with sweeps of the refresh threshold $\tau$ and observation interval $k$. Third, structure beyond flat regions: dependency-aware eviction that links a plan to the tool outputs and evidence it was built from and retains or discards them jointly, cross-region refresh when a copy of a token's content is used, region-and-phase conditioning so the same region can carry different lifetimes across agent phases, head- and layer-adaptive priorities \cite{adakv,fastgen}, region labels inferred by a lightweight classifier when no orchestrator metadata exists, and the tiered degradation path of Section~\ref*{sec:tiered}.


\end{document}